\documentclass[letterpaper, 10 pt, conference]{ieeeconf}  

\IEEEoverridecommandlockouts

\usepackage{graphicx} 
\usepackage{xcolor}
\usepackage{siunitx}
\usepackage{subcaption}
\usepackage{hyperref}
\usepackage{cleveref}
\usepackage{balance}

\title{\LARGE \bf
Seamless Integration of Tactile Sensors for Cobots
}

\author{Remko Proesmans$^{1}$ and Francis wyffels$^{1}$
\thanks{$^{1}$Remko Proesmans and Francis wyffels are with the IDLab-AIRO research lab at Ghent University -- imec, Technologiepark-Zwijnaarde
126, 9052 Zwijnaarde, Belgium {\tt\footnotesize remko.proesmans@ugent.be}}%
}

\begin{document}

\maketitle
\thispagestyle{empty}
\pagestyle{empty}

\begin{abstract}

The development of tactile sensing is expected to enhance robotic systems in handling complex objects like deformables or reflective materials.
However, readily available industrial grippers generally lack
tactile feedback, which has led researchers to develop their own tactile sensors, resulting in a wide range of sensor hardware. 
Reading data from these sensors poses an integration challenge: either external wires must be routed along the robotic arm, or a wireless processing unit has to be fixed to the robot, increasing its size.
We have developed a microcontroller-based sensor readout solution that seamlessly integrates with Robotiq grippers.
Our Arduino compatible design takes away a major part of the integration complexity of tactile sensors and can serve as a valuable accelerator of research in the field. Design files and installation instructions
can be found at \href{https://github.com/RemkoPr/airo-halberd}{https://github.com/RemkoPr/airo-halberd}.

\end{abstract}


\section{INTRODUCTION}

Tactile sensing is essential for robotic systems in dealing with complex objects in dynamic environments~\cite{billard2019}.
Clothing, for example, is difficult to handle using solely computer vision due to their infinitely large configuration space~\cite{foresti2004} and self-occlusions. 
In particular, different research groups have recently attempted a tactile approach~\cite{sunil2022, unfoldir, demiris2023} to unfold textile pieces.
Furthermore, reflective, transparent and textureless objects are a challenge for industrial robots~\cite{cendernet2023} caused by improper image segmentation and distorted stereoscopic depth images for vision-only systems.

However, commercially available industrial grippers typically do not feature tactile feedback~\cite{borras2020, birglen2018}.
The prevailing trend indicates that these grippers offer binary feedback to indicate whether they are gripping an object by monitoring their current consumption.
This feedback, combined with force torque sensing in the joints of a robotic arm, is the current extent of readily available tactile sensing.
Hence, researchers are required to acquire or develop their own tactile sensors and integrate them into the robot arm.
However, integrating tactile sensors requires careful consideration of how to power the sensors and how to communicate the data. An evident choice is to run external wiring which severely limits the movements of the robot.
In contrast, adding external batteries, power splitters or wireless readout electronics to the robot can cause self-collisions and further restrict robot movement when deployed in a constrained environment.

We present a seamless solution for the integration of tactile sensors, specifically for Robotiq grippers and Universal Robot (UR) arms, both brands that are commonly employed in literature~\cite{pieska2018, cabrera2022, bird2021, lin2019}.
By lowering the entry barrier to integrating tactile sensors to prevalent robotic grippers, we believe our design is an accelerator for many researchers in the field of robotic tactile sensing.

\begin{figure}[tpb]
\centering\vspace{4mm}
\begin{subfigure}[t]{0.5\textwidth}
    \centering
    \includegraphics[width=0.5\linewidth]{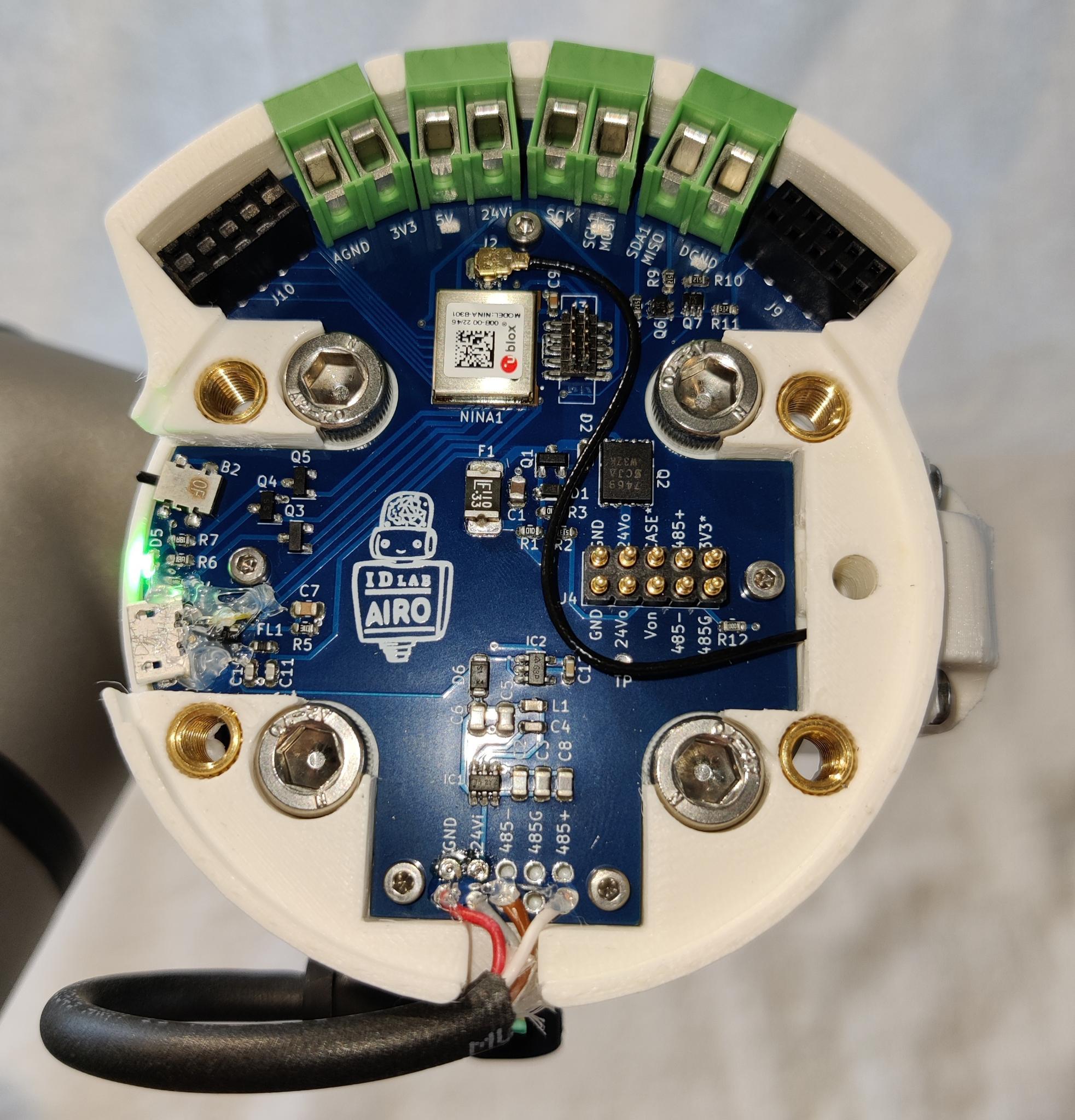}
    \caption{Internal view of custom coupling.}
    \label{fig:internal}
\end{subfigure}\vspace{2mm}\hfill
\begin{subfigure}[t]{0.2\textwidth}
    \centering
    \includegraphics[height=4.5cm]{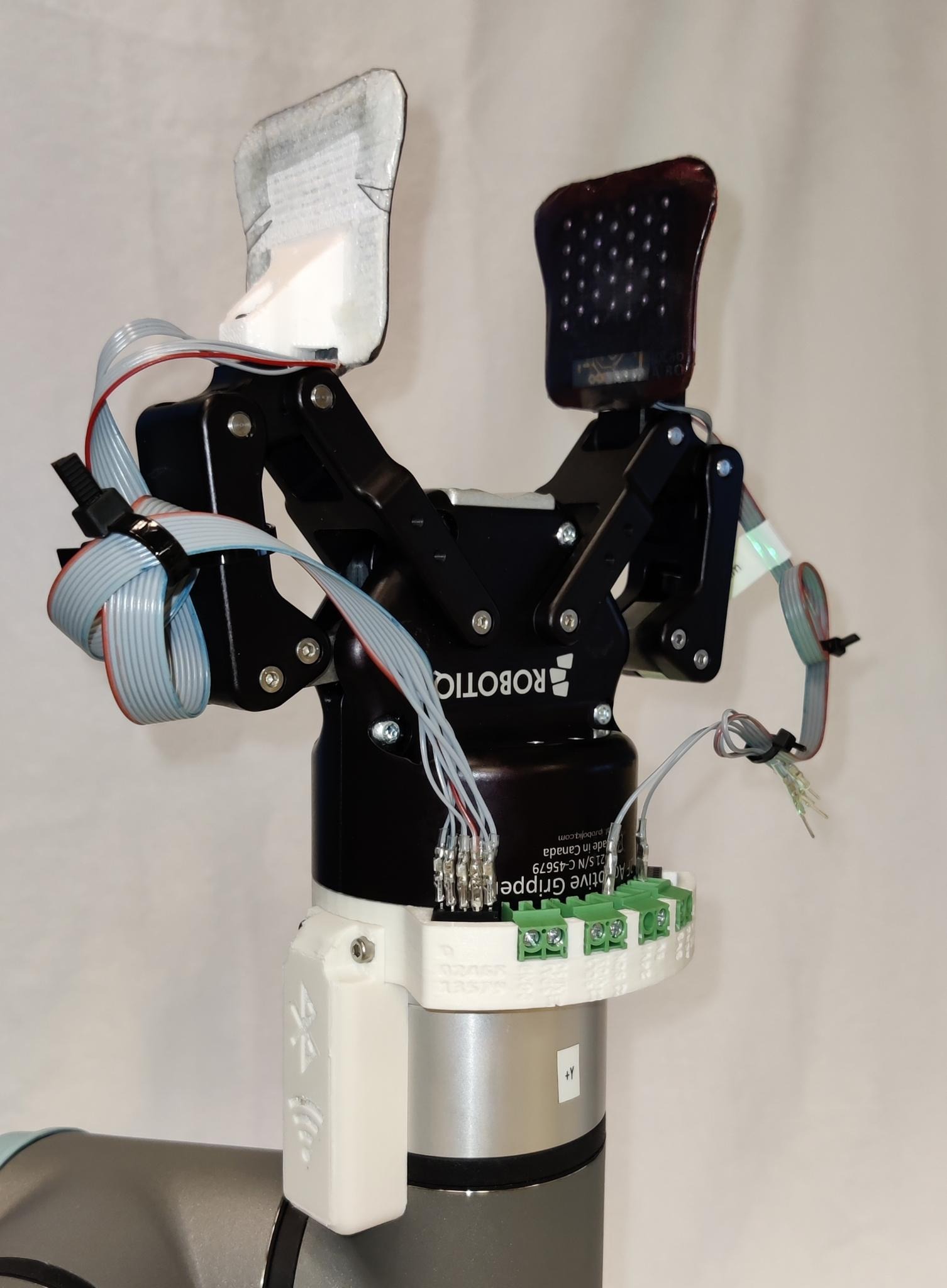}
    \caption{Robotiq 2F-85 with custom fingertips \cite{proesmans2023} mounted to UR3e arm via coupling.}
    \label{fig:integrated}
\end{subfigure}\hspace{2mm}
\begin{subfigure}[t]{0.2\textwidth}
    \centering
    \includegraphics[height=4.5cm]{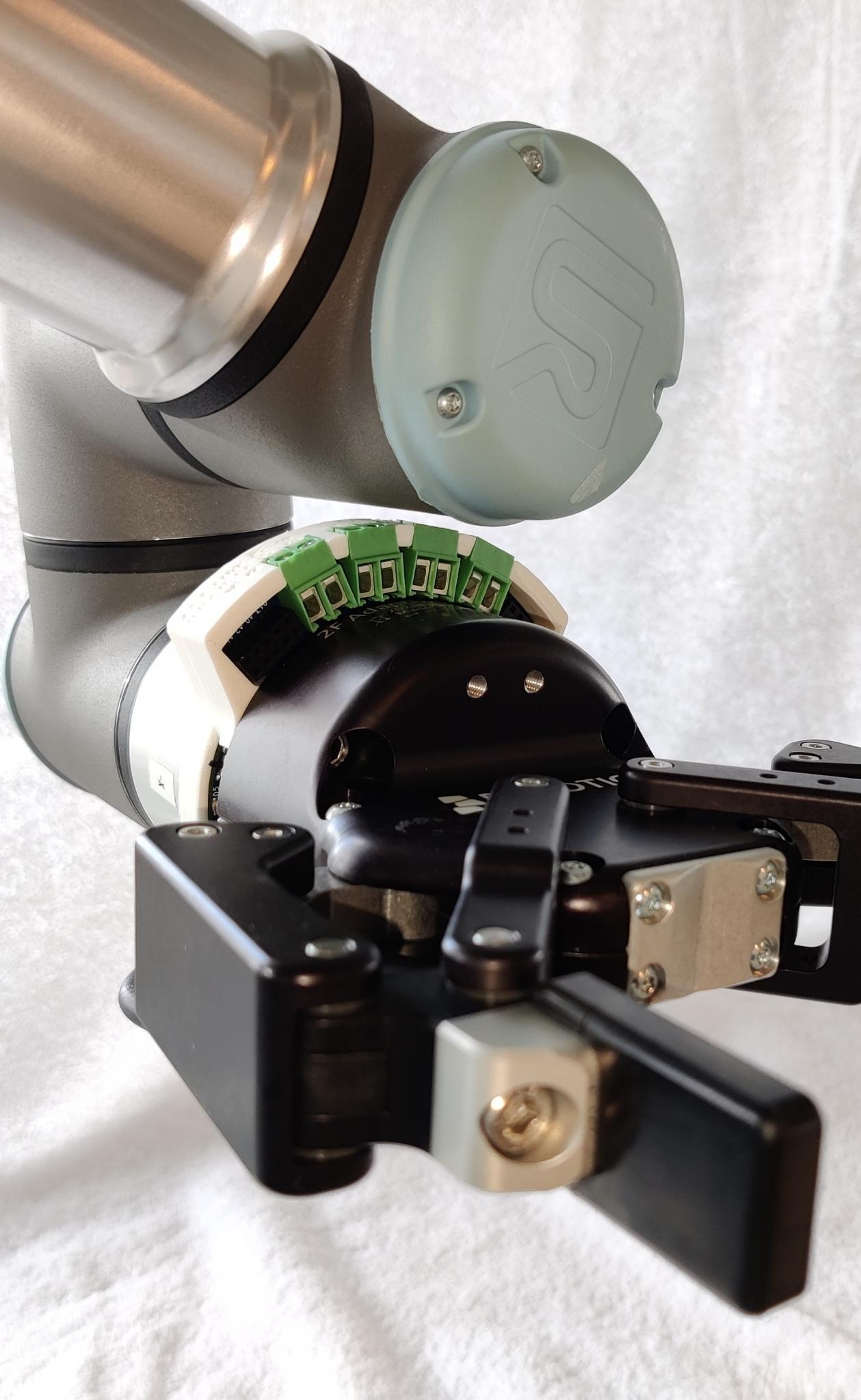}
    \caption{No self-collisions due to coupling.}
    \label{fig:self_collision}
\end{subfigure}
\caption{System integration.}
\label{fig:real_img}
\end{figure}

\section{Design}

To mount a Robotiq gripper to a cobot arm, Robotiq provides an I/O Coupling. 
This coupling is bolted to the arm, and the gripper is bolted to the coupling.
In~\cite{proesmans2023}, we presented an intermediate flange, to be placed between the Robotiq I/O Coupling and the gripper, which provided a breakout interface of the power and signal pins of the gripper.
We have now realised a coupling which completely replaces the Robotiq I/O Coupling and provides extra functionality, shown in Fig.~\ref{fig:real_img}. 

\subsection{Features}

Our coupling features the safety circuit present on the original Robotiq I/O Coupling, augmented with a Nina~B301 microcontroller unit (MCU), capable of communicating wirelessly over Bluetooth Low Energy (BLE) and Wi-Fi. 
Eight analog pins and 15 digital pins are exposed.
An SPI, UART, and two I2C interfaces are supported by a subset of the digital pins. 
The interface is fully described in the documentation on our public repository.
The board is Arduino compatible and programmable over a micro USB interface.

The following subsections describe the on-board features in detail.

\subsection{Power circuit}
\begin{figure}[tpb]
  \centering
  \includegraphics[width=\linewidth]{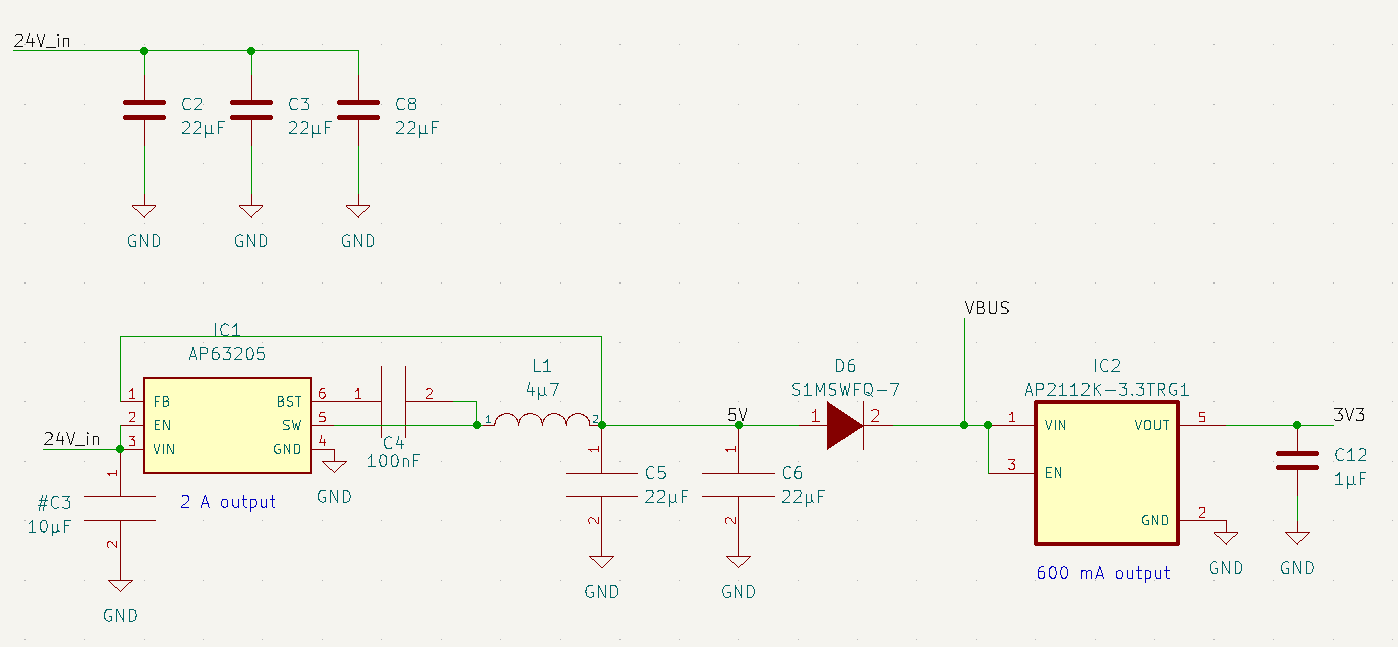}
  \caption{Power step-down circuit.}
  \label{fig:schem_power}
\end{figure}

The board either receives its power from the 24\,V UR Tool Output, like the standard Robotiq I/O Coupling, or from a micro USB connection. 
At the 24\,V input, a large bulk capacitance (Fig.~\ref{fig:schem_power} top left) prevents overcurrent faults caused by extra current drawn from the Tool Output by the MCU and sensors.
The 24\,V input voltage is stepped-down to 5\,V by a switching regulator, mitigating on-board heat dissipation, after which a linear regulator further steps down to 3.3\,V as required for the MCU.
A diode is placed between the 5\,V output and the 3.3\,V regulator input, disconnecting the 5\,V regulator when the board is powered through USB.

\begin{figure}[tpb]
  \centering
  \includegraphics[width=\linewidth]{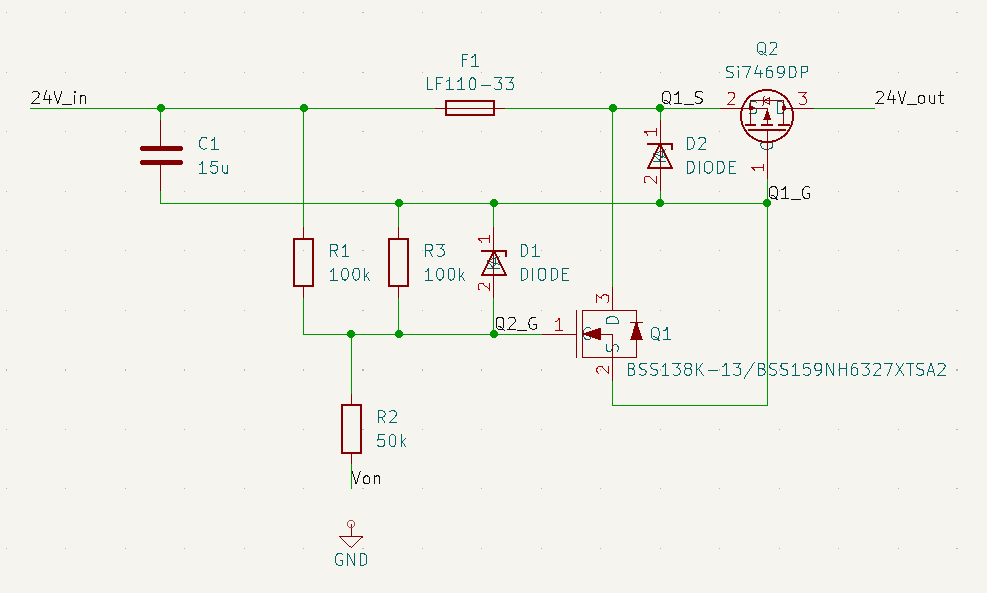}
  \caption{Power safety circuit.}
  \label{fig:schem_safety}
\end{figure}

The board also features power safety circuitry, shown in Fig.~\ref{fig:schem_safety}, inspired by the first official Robotiq I/O Coupling.
The 24\,V input of the coupling is passed through a 2\,mA resettable fuse, as well as the channel of a P-channel MOSFET (Q2), before reaching the gripper.
Q2 is biased by Zener diode D2 and resistors R3 and R2.
The connection of R2 to ground is, however, only made when the gripper is physically attached to the coupling.
Hence, when the gripper is detached, D2 cannot bias Q2 and the 24\,V output of the coupling is detached from the 24\,V input.
Q1 is a depletion-type N-channel MOSFET, which ensures that the gate of Q2 is quickly discharged upon detachment of the gripper.


\subsection{Microcontroller circuit} \label{s:mcu}
\begin{figure}[tpb]
  \centering
  \includegraphics[width=\linewidth]{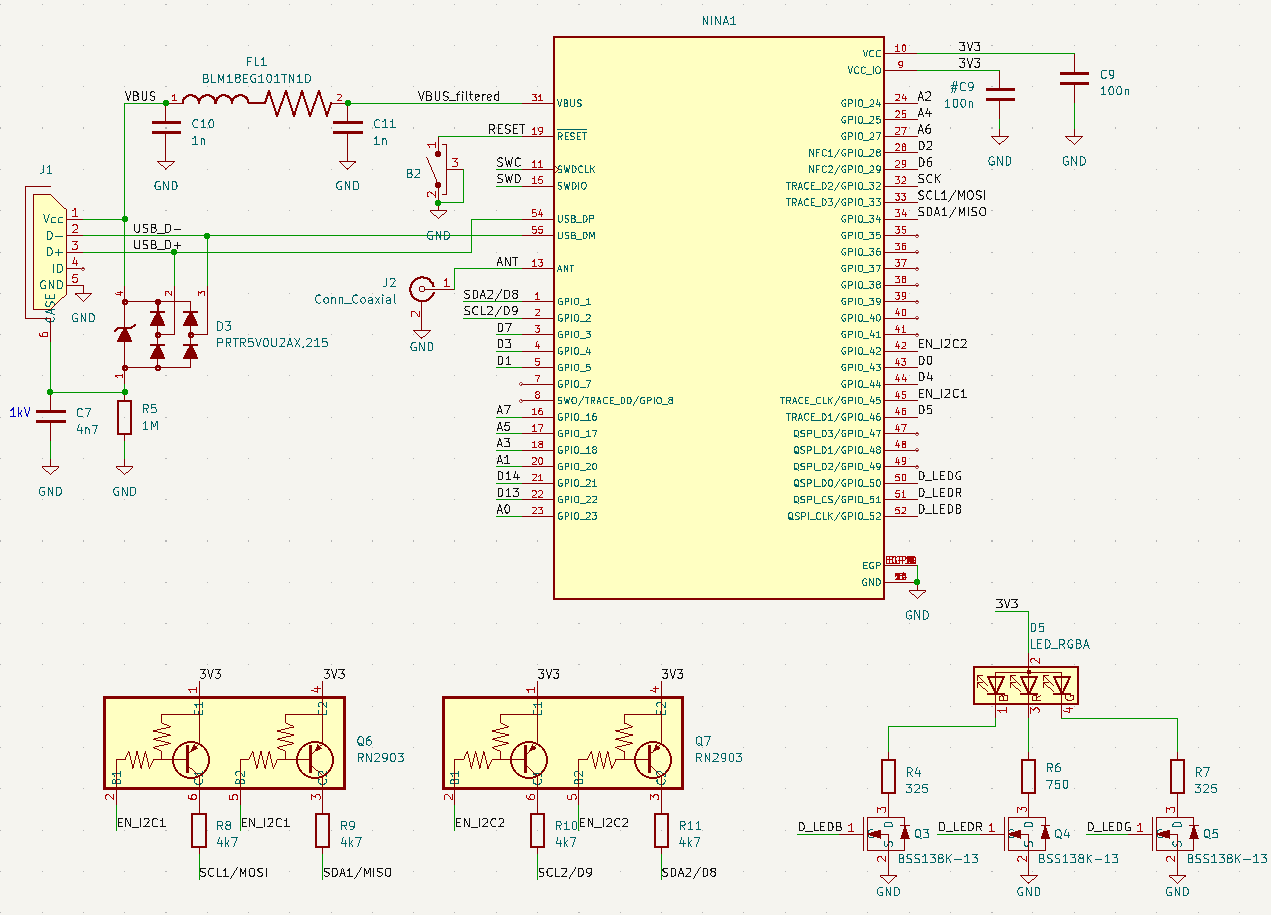}
  \caption{Microcontroller circuit.}
  \label{fig:schem_nina}
\end{figure}

The on-board MCU is the u-blox\textsuperscript{\tiny\textregistered} Nina~B301 (Fig.~\ref{fig:schem_nina}, top centre).
We base the design of our PCB on the tried-and-tested Arduino Nano 33 BLE, which uses the Nina~B306 with integrated PCB antenna. 
We opted for the Nina~B301 with an external 2.45\,GHz antenna, to be attached on the exterior of the flange, as we suspect that the radiation characteristics of the on-chip antenna would strongly deteriorate when placed in the coupling between the robotic arm and the gripper.

The micro USB filter and safety circuitry on our board, on the left side of Fig.~\ref{fig:schem_nina}, is directly taken from the Arduino Nano 33 BLE design files.

Furthermore, we have improved upon the Arduino design by ensuring that the pull-up resistors for the I2C interfaces (bottom left of Fig.~\ref{fig:schem_nina}) are programmatically detachable. 
This is unlike the Arduino design where the I2C pins remain interconnected through a series connection of their pull-up resistors when deployed as general-purpose pins.

Lastly, a red, green and blue light-emitting diode (RGB LED) is provided as a programmable status indicator.

\subsection{Casing}
The casings for the microcontroller board and the 2.45\,GHz antenna are designed in SolidWorks and 3D printed in PLA using a Prusa~i3~MK3.
The dimensions of our coupling are chosen such that the risk of self-collisions is equivalent to the default Robotiq I/O Coupling.
Additionally, the height of our coupling is exactly the same as that of the original, so that the couplings can be swapped without having to redefine the tool centre point of the robot. 
To ensure both durability and robustness, we employ heat-set inserts for the bolt connections to mitigate wear and bolster mechanical integrity.

\section{Software}
As the Nina~B301 present on our board is nearly identical to the Nina~B306 of the Arduino Nano 33 BLE, it is possible to use the Arduino bootloader and core files to program our board.
To this end, we have provided custom Arduino board files which are readily installable through the Arduino IDE.
When the board is programmed for the first time, the Arduino bootloader has to be flashed by using the Serial Wire Debug protocol, for which a 10-pin header and additional solder pads on the bottom of the board are provided.
Detailed information on this is provided in the public repository.
Once the Arduino bootloader is flashed, the board can also be programmed from the Arduino IDE over the micro USB interface.

\section{CONCLUSION}
Researching and employing the potential of tactile sensing in robotics requires careful consideration of how to power the sensors and how data is communicated to a remote host.
External wiring can strongly constrain the movements of the robot, and external mounting of wireless components can cause collisions.
We have provided a seamless solution to integrate tactile sensors on Robotiq grippers and Universal Robot arms.
Robotiq also provides I/O Couplings for several other cobot brands such as Fanuc~CRX and Yaskawa.
Evaluating the extent to which our coupling must be redesigned to fit other cobot brands is left to future work.
By providing this augmentation of current state-of-the-art hardware, we hope to accelerate research in tactile sensing and bring robotic manipulation a step closer to human performance for complex objects such as deformables and metallics.

\section{ACKNOWLEDGEMENT}
Remko Proesmans is a predoctoral fellow of the Research Foundation Flanders (FWO) under grant agreement no. 1S15923N. 
This work was also partially supported by the euROBIn Project (EU grant number 101070596).

\addtolength{\textheight}{-12cm}   





\balance

\end{document}